\documentclass{article}
\usepackage{natbib} 
\usepackage{microtype}
\usepackage{graphicx}
\usepackage{subfigure}
\usepackage{booktabs} 
\usepackage{lipsum} 
\usepackage{pifont} 
\usepackage{hyperref}
\usepackage{algorithm}
\usepackage{algpseudocode}
\usepackage{multirow}

\usepackage[arxiv]{icml2025}
\usepackage{amsmath}
\usepackage{amssymb}
\usepackage{mathtools}
\usepackage{amsthm}

\usepackage{soul}
\usepackage{xcolor}

\algtext*{EndIf}     
\algtext*{EndFor}     
\algtext*{EndFunction} 

\newcommand\eat[1]{}
\setstcolor{red}

\usepackage[capitalize,noabbrev]{cleveref}

\theoremstyle{plain}

\theoremstyle{definition}

\theoremstyle{remark}

\usepackage[textsize=tiny]{todonotes}

\icmltitlerunning{A $\lambda$-Compass for AIGC Provenance}

\begin{document}

\twocolumn[
\icmltitle{Lost in Edits? A \texorpdfstring{$\lambda$}{lambda}-Compass for AIGC Provenance}



\icmlsetsymbol{equal}{*}
\begin{icmlauthorlist}
\icmlauthor{Wenhao You}{uofw}
\icmlauthor{Bryan Hooi}{nus}
\icmlauthor{Yiwei Wang}{ucm}
\icmlauthor{Euijin Choo}{uofa} 
\icmlauthor{Ming-Hsuan Yang}{ucm}
\icmlauthor{Junsong Yuan}{sunyb}
\icmlauthor{Zi Huang}{uq}
\icmlauthor{Yujun Cai}{uq}
\end{icmlauthorlist}
\icmlaffiliation{uofw}{Department of Electrical and Computer Engineering, University of Waterloo}
\icmlaffiliation{uofa}{Department of Computing Science, University of Alberta}
\icmlaffiliation{nus}{Department of Computer Science, National University of Singapore}
\icmlaffiliation{ucm}{Department of Computer Science and Engineering, University of California, Merced}
\icmlaffiliation{sunyb}{Department of Computer Science and Engineering, University at Buffalo, State University of New York}
\icmlaffiliation{uq}{School of Electrical Engineering and Computer Science, University of Queensland}

\icmlcorrespondingauthor{Wenhao You}{w22you@uwaterloo.ca}
\icmlkeywords{Machine Learning}

\vskip 0.3in
]



\printAffiliationsAndNotice{}  

\begin{abstract}
Recent advancements in diffusion models have driven the growth of text-guided image editing tools, enabling precise and iterative modifications of synthesized content. However, as these tools become increasingly accessible, they also introduce significant risks of misuse, emphasizing the critical need for robust attribution methods to ensure content authenticity and traceability. Despite the creative potential of such tools, they pose significant challenges for attribution, particularly in adversarial settings where edits can be layered to obscure an image’s origins. We propose \textsc{LambdaTracer}, a novel latent-space attribution method that robustly identifies and differentiates authentic outputs from manipulated ones without requiring any modifications to generative or editing pipelines. By adaptively calibrating reconstruction losses, \textsc{LambdaTracer} remains effective across diverse iterative editing processes, whether automated through text-guided editing tools such as InstructPix2Pix and ControlNet or performed manually with editing software such as Adobe Photoshop. Extensive experiments reveal that our method consistently outperforms baseline approaches in distinguishing maliciously edited images, providing a practical solution to safeguard ownership, creativity, and credibility in the open, fast-evolving AI ecosystems.
\end{abstract}

\section{Introduction}~\label{introduction}\vspace{-2pt} 
Recent advancements in diffusion models, such as Stable Diffusion~\cite{rombach2022high}, have significantly propelled the field of image generation. Concurrently, text-guided image editing models like InstructPix2Pix~\cite{brooks2023instructpix2pix} and ControlNet~\cite{zhang2023adding,zhao2024uni,li2025controlnet} have emerged, enabling modifications based on user instructions. Commercial services, including DALL-E 3~\cite{betker2023improving} from OpenAI, Anytext~\cite{tuo2023anytext} from Alibaba, and Photoshop from Adobe, further democratize access through user-friendly APIs and tools. While these advancements empower creative applications by lowering the barriers to image creation and editing, they simultaneously introduce opportunities for misuse by malicious actors~\cite{verge2024appleai,verge2024googleai,ke2025detection}. 
Specifically, malicious individuals may falsely claim AI-edited photographs or artworks as their original creations, posing significant intellectual property rights and authenticity verification challenges. Moreover, judicial rulings in several countries have recognized that model-generated images meeting originality criteria are eligible for copyright protection\cite{he2024ai}, sparking widespread debate~\cite{ren2024copyright,ducru2024ai,gaffar2024copyright}. These challenges underscore the critical importance of addressing copyright issues associated with synthesized content and highlight the urgent need for methods to detect and trace the origin of images, identify the underlying generation models, and determine whether someone has edited the images.
\begin{figure}[!t]
    \centering
    \includegraphics[width=0.45\textwidth]{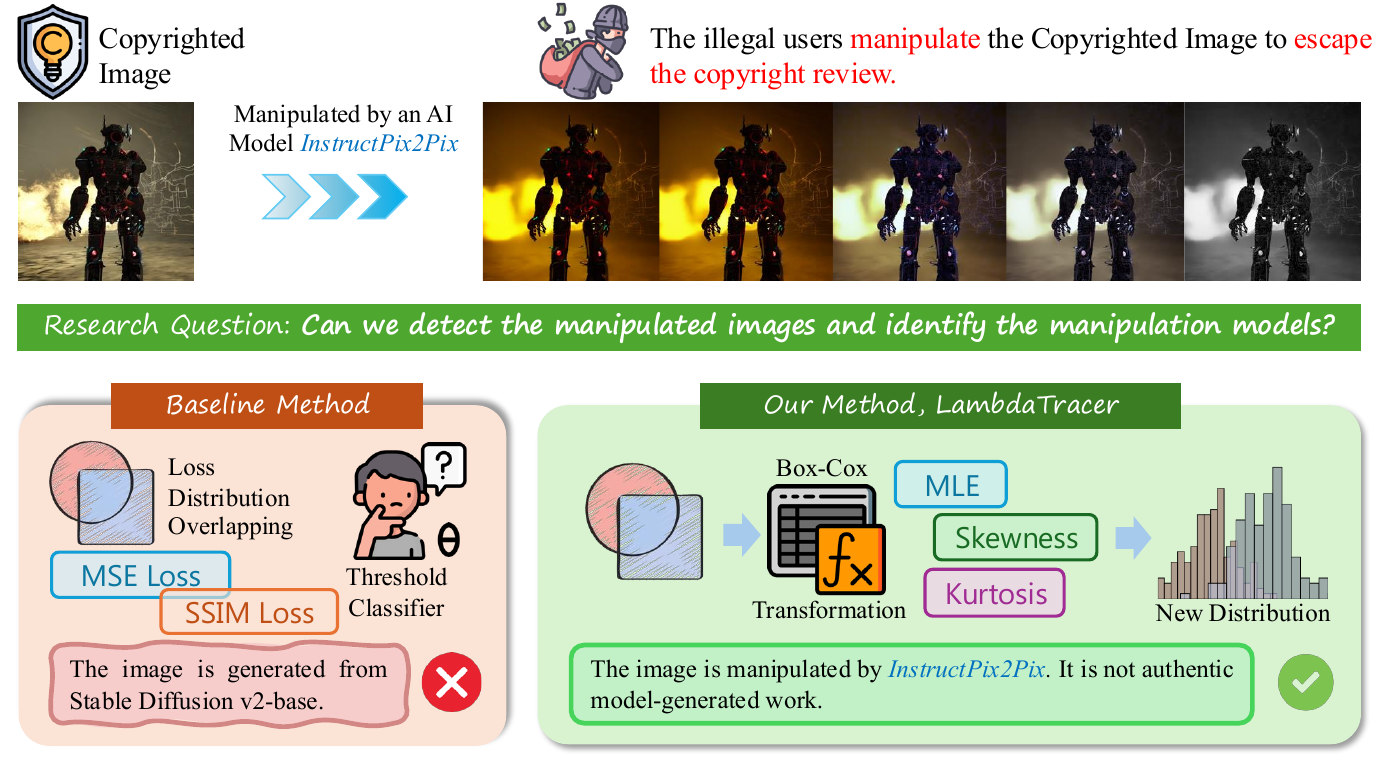}
    \vspace{-10pt} 
    \caption{An example of iterative text-guided editing InstructPix2Pix~\cite{brooks2023instructpix2pix} on model-generated images. The baseline \textsc{LatentTracer}~\cite{wang2024trace} struggles with cumulative perturbations, causing provenance inconsistencies. Our method, \textsc{LambdaTracer}, ensures robust tracing, effectively distinguishing authentic and manipulated content.}
    \label{fig:6imgs}
    \vspace{-10pt} 
\end{figure}

Existing methods for detecting the source of generated images can be broadly categorized into two approaches. The first, \emph{embedding-based detection methods}, involve injecting fingerprint information during training~\cite{tancik2020stegastamp,yu2021artificial} or modifying model architectures to embed detectable fingerprints into generated images~\cite{yu2020responsible,jeong2022fingerprintnet,sinitsa2024deep}. However, embedding-based methods are often impractical in open environments due to the infeasibility of universally enforcing watermark embedding during training or generation. The second approach, \emph{latent space reverse-engineering detection methods}, attempts to trace the origin of an image by working backward. Specifically, these methods evaluate whether an image can be accurately recreated by feeding it back into the model and finding a representation in the model’s internal \emph{latent space} that closely matches the original image~\cite{asnani2023reverse,wang2023alteration,wang2024did,wang2024trace}. This tests whether an image aligns with a model’s internal structure, enabling direct source attribution. However, they fail against more sophisticated adversaries who iteratively manipulate images via text-guided editing models, leading to provenance inconsistencies (Figure~\ref{fig:6imgs}). In contrast, our method, \textsc{LambdaTracer}, effectively traces both original and edited images. 
%
%
                     
In this paper, we address these challenges by proposing \textsc{LambdaTracer}, a novel origin attribution method specifically designed to tackle the complexities introduced by text-guided image editing models. Our key contributions are: \textbf{\ding{172} Robust and versatile origin attribution:} We propose \textsc{LambdaTracer}, an alteration-free, inversion-based framework that effectively handles challenges from text-guided models (e.g., InstructPix2Pix~\cite{brooks2023instructpix2pix}, ControlNet~\cite{zhang2023adding,zhao2024uni,li2025controlnet}) and image editing tools (e.g., Adobe Photoshop), ensuring reliability in open and adversarial settings. \textbf{\ding{173} First systematic study on text-guided editing in attribution tasks:} We present the first comprehensive analysis of the impact of text-guided editing methods on image attribution, particularly for artistic content. By considering the potential misuse by adversaries, we provide a research context that closely reflects real-world challenges.   
\textbf{\ding{174} Flexible loss transformation:} We introduce a flexible loss transformation approach based on an improved Box-Cox Transformation~\cite{box1964analysis}. This method significantly enhances the separability of overlapping distributions, improving attribution accuracy across diverse scenarios.  
\textbf{\ding{175} State-of-the-art performance:} Our method outperforms baselines in adversarial settings, achieving the best attribution accuracy across both generated and edited images. 

\section{Related Work}\label{sec:related}\vspace{-2pt} 
\textbf{Stable Diffusion Generative Models.} Stable Diffusion~\cite{rombach2022high} has emerged as a major advancement in generative models, surpassing traditional models such as GANs~\cite{goodfellow2014generative,radford2015unsupervised,karras2019style} and VAEs~\cite{kingma2013auto,tolstikhin2017wasserstein} in terms of computational efficiency, scalability, and output quality. Stable Diffusion employs a denoising diffusion process within a low-dimensional latent space, enhancing robustness and significantly reducing memory and computational requirements compared to pixel-space methods ~\cite{ho2020denoising,song2020score} while maintaining high-quality image generation. Across different versions, Stable Diffusion has introduced key improvements in model architecture, training datasets, and noise-handling techniques, resulting in higher resolution outputs and improved alignment with textual prompts~\cite{rombach2022high}. These advancements have transformed creative applications, enabling breakthroughs in art generation, design prototyping, and visual storytelling~\cite{han2023design,wu2023not,wang2024diffusion}.

\textbf{Text-guided Image Editing Methods.} Text-guided image editing methods~\cite{li2020manigan,choi2023custom,wang2023imagen,ravi2023preditor} have seen significant progress with the advent of conditional diffusion models~\cite{nichol2021improved,dhariwal2021diffusion}, which enhance controllability and flexibility in generating or editing images. InstructPix2Pix~\cite{brooks2023instructpix2pix} is a conditional diffusion model designed explicitly for text-guided image editing tasks. It takes an input image and a natural language instruction as conditions, enabling precise edits to the image. By fine-tuning instruction-image pairs, it excels in tasks like style transfer and scene adjustments. Similarly, ControlNet~\cite{zhang2023adding,zhao2024uni,li2025controlnet} extends the framework of conditional diffusion models by incorporating structural guidance inputs such as edge maps, pose data or segmentation maps. ControlNet preserves spatial coherence and structural integrity while enabling high-fidelity edits based on textual and structural conditions. While these models represent state-of-the-art advancements in text-guided image editing, they also introduce challenges for origin attribution. Specifically, iterative use of these models can produce images with complex modification histories, complicating efforts to trace the source or the sequence of edits. Addressing these challenges is critical for ensuring content traceability and accountability.

\textbf{Generative Content Provenance Methods.} Current methods for tracing the origins of generated images can be broadly divided into two main approaches:
Embedding-based detection and latent space reverse-engineering. Embedding-based detection methods~\cite{tancik2020stegastamp,yu2021artificial,yu2020responsible,jeong2022fingerprintnet,sinitsa2024deep}, often referred to as fingerprinting or watermarking techniques, embed unique identifiers into models during training to trace the origin of model-generated images. These identifiers can later be detected in generated outputs to verify image provenance. While effective, they require training modifications, making them impractical for pre-trained or open-source models, and are vulnerable to adversarial removal or forgery~\cite{xu2020adversarial,sun2021detect}. Additionally, embedding increases complexity and frequently affects image quality.

Latent space reverse-engineering methods~\cite{creswell2018inverting,wang2023alteration,wang2024did,wang2024trace} trace the origin of generated images by 
optimizing latent vectors to minimize reconstruction loss, thereby avoiding model modifications. However, these methods are limited to original model-generated images and 
fail against cumulative perturbations from iterative text-guided editing models~\cite{brooks2023instructpix2pix,zhang2023adding,zhao2024uni,li2025controlnet}, limiting effectiveness in open environments. To address this gap, we propose \textsc{LambdaTracer}, a more robust latent space reverse-engineering approach that can both attribute origins and detect iterative modifications, substantially expanding the scope of provenance tracing in real-world scenarios.

\section{Latent Distribution Analysis }\label{sec:distribution_analysis}
\vspace{-2pt} 
As discussed in \S~\ref{sec:related}, embedding-based methods fail in open-source contexts due to their reliance on model modifications. In contrast, latent space reverse-engineering methods are compromised by the cumulative distortions from iterative editing. 
Among these, threshold-based classification~\cite{wang2023alteration} is the most effective. However, this approach struggles in scenarios involving perturbed or edited images due to significant overlap in the reconstruction loss distributions between different categories, such as pristine model-generated images and modified outputs. To illustrate how repeated modifications blur the line between generated and edited outputs, we analyze the distribution of reconstruction losses in this section, providing both empirical and theoretical insights into this overlap. 

\subsection{Quantitative Analysis of Distribution} 
We examine the reconstructed Mean Squared Error (MSE) loss distributions across different image categories by calculating the overlap between their probability density functions (PDF), estimated using Gaussian Kernel Density Estimation (KDE)~\cite{davis2011remarks}. Equation~\ref{equ:kde} illustrates the KDE formulation, which provides smooth estimates of the PDFs for the given data points. Once the PDFs, $f_1$ and $f_2$, are estimated, Equation~\ref{equ:overlap} illustrates the overlap calculation, where higher overlap values indicate reduced separability between the distributions. 
\begin{align}
f(x) &= \frac{1}{n h \sqrt{2\pi}} \sum_{i=1}^{n} \exp\left( -\frac{(x - x_i)^2}{2 h^2} \right) \label{equ:kde} \\
\mathcal{O}(x) &= \int_{-\infty}^{\infty} \min\bigl(f_1(x),f_2(x)\bigr) dx \label{equ:overlap} 
\end{align}
where $ n $ is the number of data points, $ h $ is the bandwidth, and $ x_i $ represents each data point.

\begin{figure}[t]
    \centering
    \includegraphics[width=0.45\textwidth] {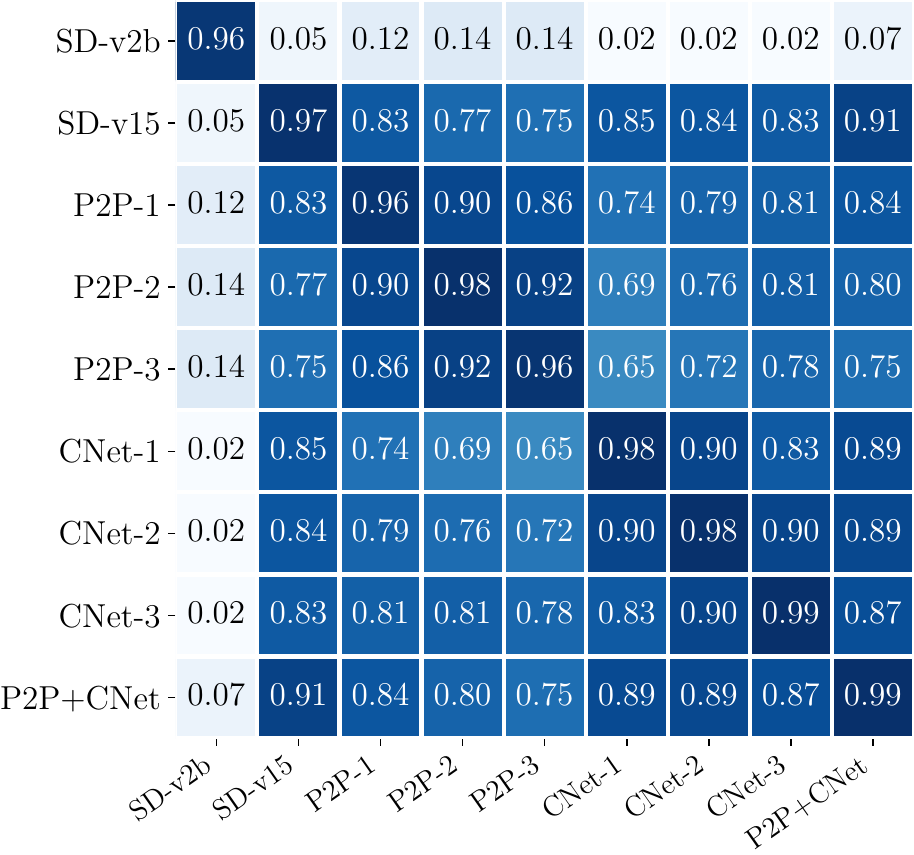}
    \vspace{-10pt} 
   \caption{Heatmap of PDF overlap in loss distributions across image categories. Rows and columns correspond to different image types: originally generated, single-edited, and iteratively edited images. P2P denotes the InstructPix2Pix~\cite{brooks2023instructpix2pix}, while CNet represents the ControlNet~\cite{li2025controlnet}. The suffix (e.g., P2P-1) indicates that the corresponding model was applied $n$ times iteratively. Each cell, ranging from $0$ to $1$, quantifies the degree of overlap, where higher values indicate greater similarity in loss distributions, and lower values represent better separability.}
    \label{fig:heatmap_ori}
    \vspace{-10pt}
\end{figure}

Fig.~\ref{fig:heatmap_ori} visualizes the overlap across image types using a heatmap, where darker regions indicate better separability (lower overlap) and lighter regions signify greater similarity (higher overlap). We utilize Stable Diffusion v2-base (SD v2-base) and Stable Diffusion v1-5 (SD v1-5) ~\cite{rombach2022high} to create images without text-guided editing; we apply InstructPix2Pix~\cite{brooks2023instructpix2pix} and ControlNet~\cite{zhang2023adding,zhao2024uni,li2025controlnet} with three iterative edits to create text-guided edited images. Details on models, prompts, and editing processes are provided in \S~\ref{subsection:experiment_setup} and Appendix~\ref{appendix:edited_prompts}.
As shown in Fig.~\ref{fig:heatmap_ori}, the overlap between Stable Diffusion v2-base and Stable Diffusion v1-5 remains minimal, indicating strong separability for originally generated images. However, text-guided editing, especially with iterative modifications, increases distributional overlap. This effect is particularly evident with InstructPix2Pix and ControlNet, where three iterative edits lead to significant convergence with their original counterparts, emphasizing the cumulative impact of editing on attribution tasks. 

\subsection{Theoretical Analysis of Distribution}
We can formalize the accumulation of editing perturbations in Equation~\ref{equ:theo_analysis}.
\begin{align}
\label{equ:theo_analysis}
z_n = z_0 + \sum_{i=1}^{n} f\bigl(z_{i-1}, \theta_i\bigr),
\end{align}
where $z_0$ represents the initial latent vector and $f\bigl(z_{i-1}, \theta_i\bigr)$ denotes the nonlinear transformation included by the editing model at the $i$-th step, parameterized by $\theta_i$. 

The equation follows from an iterative formulation, where each editing step applies a transformation conditioned on the previous latent state, modelling the accumulated modifications in a structured way. Each edit perturbs the latent representation non-linearly, leading to a gradual distributional drift. The term $\sum_{i=1}^{n} f\bigl(z_{i-1}, \theta_i\bigr)$ directly captures how small perturbations accumulate over multiple edits, leading to progressive drift. Over multiple editing steps, this drift may push $z_n$ away from the original data manifold, causing it to overlap with the distributions of other categories. This phenomenon explains why distinguishing between original and edited content becomes increasingly difficult: as modifications accumulate, the latent space regions that initially correspond to distinct classes blur together. From a discriminative perspective, this drift directly impacts classification tasks. If $z_n$ deviates significantly from the original distribution but aligns with other categories, a classifier trained on the original data distribution may struggle to differentiate manipulated content. This provides intuitive reasoning for why standard reconstruction loss comparison is insufficient: reconstruction losses typically measure pixel-wise or feature-level similarity without explicitly accounting for how edits influence class separability in the latent space.

\subsection{Key Insights and Motivation}
These quantitative and theoretical analyses reveal two key insights. First, iterative editing introduces cumulative distortions that significantly alter the distribution of reconstruction losses, with overlap rates increasing by up to 86\% after multiple edits (as shown in Fig.~\ref{fig:heatmap_ori}). Second, the non-linear nature of these distortions suggests that conventional linear approaches would be insufficient for reliable attribution. These findings highlight the need for an adaptive approach that can effectively handle varying degrees of distribution overlap while preserving discriminative features.

\section{Methodology}
As advanced generative models and easily accessible editing tools amplify threats to content authenticity and intellectual property, a robust attribution framework for real-world scenarios involving iterative text-guided editing is critically needed. We first provide a formal definition of our attribution task (\S~\ref{subsection:problem_formulation}).
Next, we provide an overview of our proposed \textsc{LambdaTracer} system (\S~\ref{sec:overview_lam}), followed by a detailed explanation of the transformation design (\S~\ref{sec:transformation_design}) and a dynamic parameter selection strategy (\S~\ref{sec:lambda_selection}). We describe our lightweight supervised classification mechanism (\S~\ref{subsection:svm_classification}).

\subsection{Problem Formulation}~\label{subsection:problem_formulation}
\textbf{Attribution Task.} Our goal is to attribute a target image $\mathcal{I}_t$, which is directly generated by a generative model ($\mathcal{M}_g$) or modified by an editing model ($\mathcal{M}_e$). This involves distinguishing between images directly generated by a base generative model (e.g., Stable Diffusion~\cite{rombach2022high}) and those modified iteratively using text-guided editing tools such as InstructPix2Pix~\cite{brooks2023instructpix2pix} or ControlNet~\cite{zhang2023adding}. Formally, this task can be framed as a binary classification problem, where the objective is to design a mapping function $\mathcal{F}$ that takes as input the target image $\mathcal{I}_t$ and outputs a label $y \in \{1, 2\}$. The mapping function can be expressed as $\mathcal{F} : \mathcal{I}_t \mapsto y$.
\begin{figure*}[h]
    \centering
    \includegraphics[width=1\textwidth]{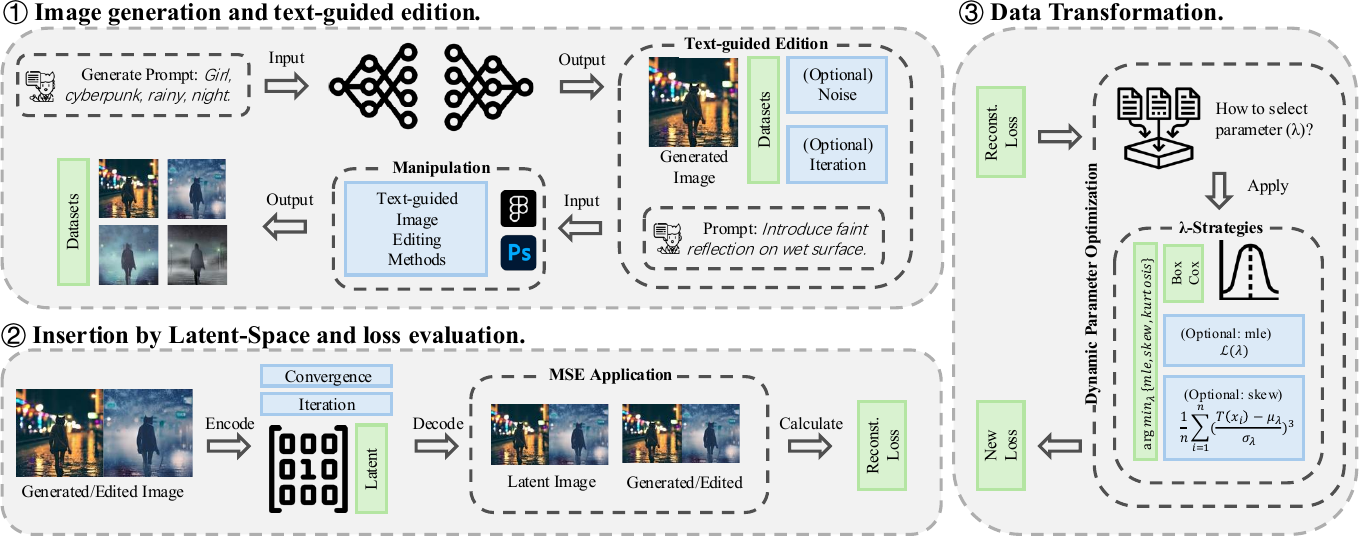}
    \vspace{-15pt} 
    \caption{The proposed pipeline consists of three steps: \ding{172} Image generation and edition by using stable diffusion models and text-guided editing methods respectively; \ding{173} Latent-space insertion and reconstruction loss evaluation via mean square error (MSE); \ding{174} Data Transformation and Dynamic $\lambda$-selection strategies, including maximum likelihood, skewness minimization, and kurtosis minimization.}
    \label{fig:general_process}
    \vspace{-10pt} 
\end{figure*}
\subsection{Overview of \textsc{LambdaTracer}}
\label{sec:overview_lam}
We propose \textsc{LambdaTracer}, a robust attribution approach tailored to scenarios where an image may have undergone multiple text-guided edits or other manipulations. By combining a carefully chosen loss transformation with dynamic parameter selection, our method aims to reduce the overlap of reconstruction-error distributions for generated and iteratively edited images and maintain consistent performance even under adversarial manipulations. The framework operates without embedding watermarks or modifying source models, making it viable in real-world, open environments. Fig.~\ref{fig:general_process} illustrates the general process of \textsc{LambdaTracer}. 

\subsection{Transformation Design}
\label{sec:transformation_design}
To handle varying degrees of distribution overlap, it is important to choose a proper transformation.

\textbf{Comparison of Transformations.} We evaluate several transformations, including Z-score standardization, logarithmic transformation, exponential transformation, and power transformation (see Appendix~\ref{appendix:transformations} for more details). 
We find that Z-score standardization, while effective for normalizing data scales, fails to address the non-linear distortions introduced by iterative edits. Logarithmic transformation compresses the right tail of skewed distributions but offers limited flexibility in capturing subtle differences across diverse editing intensities. Although theoretically capable of amplifying differences, exponential transformation tends to exaggerate outliers and noise. Power transformation provides greater flexibility through its power parameter but lacks the adaptability to handle varying editing intensities effectively. \looseness=-1

These approaches suffer from a fundamental limitation: they apply a fixed transformation regardless of the underlying distribution characteristics. This inflexibility makes them inadequate for handling the complex distortions introduced by repeated edits, often degrading key attribution features. To address these limitations, we propose using the Box-Cox transformation~\cite{box1964analysis}, which offers adaptive flexibility through its tunable parameter $\lambda$.

\textbf{Box-Cox Transformation.} We argue that Box-Cox transformation~\cite{box1964analysis} is the most suitable choice for our task for the following reasons. First, it addresses skewness in the loss distributions by compressing heavily skewed tails and pushing the data toward a more symmetric shape. This property is particularly crucial for iterative edits, where repeated modifications introduce cumulative distortions. Second, its tunable parameter $\lambda$ provides flexible control over the transformation intensity, allowing adaptation to diverse editing patterns. Finally, the transformation preserves the monotonic ordering of the data, ensuring that critical distinctions in reconstruction errors are not obscured. Equation~\ref{equ:boxcox} presents the general form of the Box-Cox transformation, where $\lambda$ governs the form and strength of the power-based adjustment. Applying Box-Cox to the reconstructed loss values substantially improved the separability of generated versus edited images, particularly in highly adversarial settings.
\begin{align}~\label{equ:boxcox}
T_\lambda(y_i)  =
\begin{cases}
\frac{y_i^\lambda - 1}{\lambda}, & \text{if } \lambda \neq 0, \\
\ln(y_i), & \text{if } \lambda = 0,
\end{cases}
\end{align}
\text{where } $y_i$ is the $i$-th original data point, and $\lambda$ is a tunable parameter that controls the strength and form of the transformation.

\subsection{$\lambda$ Selection Strategies}
\label{sec:lambda_selection}

One of the key challenges in applying the Box-Cox transformation is selecting the optimal value of $\lambda$. An effective strategy must balance adaptability to various data distributions and computational efficiency. To address this, we propose three approaches: Maximum Likelihood Estimation (MLE), Skewness Minimization, and Kurtosis Minimization, each targeting specific aspects of distributional distortion. The core formulas for these strategies and detailed pseudo-codes are given in Appendix~\ref{appendix:formula} and Appendix~\ref{appendix:algo}, respectively.

\textbf{Strategy 1: Maximum Likelihood Estimation (MLE).}
This strategy identifies $\lambda$ by maximizing the likelihood $ \mathcal{L}(\lambda)$ of the transformed data under a Gaussian assumption. Formally, we solve $\lambda_{\text{MLE}} = \arg\max_\lambda \mathcal{L}(\lambda)$. The high-level idea is that by favouring a transformation approximating normality, we achieve more stable and separable distributions, which is crucial for attribution tasks. \textit{Algorithm~\ref{alg:boxcox_mle}} illustrates how we iterate over a discrete set of candidate $\lambda $values, compute their likelihood scores, and select the one yielding the highest $\mathcal{L}$. This approach often proves effective when data exhibit complex, mixed distortions due to multiple or varied editing steps.
\begin{algorithm}[H]
    \caption{MLE for $\lambda$ Selection}
    \label{alg:boxcox_mle}
    \textbf{Input:} Loss Values $x$ \\
    \textbf{Output:} Transformed Data $t$, Optimal Parameter $\lambda^*$
    \begin{algorithmic}[1]
        \Function{BoxCox-MLE}{$x$}
            \State $x \gets \{x_i + 10^{-8} \mid x_i \in x\}$
            \State $\Lambda \gets \{\lambda_1, \lambda_2, \dots, \lambda_n\}$
            \State $\lambda^* \gets \text{None}$
            \State $\mathcal{L}_{\text{max}} \gets -\infty$
            \For{$\lambda \in \Lambda$}
                \State $\mathcal{L}(\lambda) \gets \text{Likelihood of transformed } x$
                \If{$\mathcal{L}(\lambda) > \mathcal{L}_{\text{max}}$}
                    \State $\mathcal{L}_{\text{max}} \gets \mathcal{L}(\lambda)$
                    \State $\lambda^* \gets \lambda$
                \EndIf
            \EndFor
            \State $t \gets \text{BoxCox}(x, \lambda^*)$
            \State \Return $t, \lambda^*$
        \EndFunction
    \end{algorithmic}
\end{algorithm}
\vspace{-15pt} 

\textbf{Strategy 2: Skewness Minimization.}
The second strategy focuses on reducing asymmetry in the transformed data by minimizing the absolute skewness $\left|S(\lambda)\right|$, where $S(\lambda)$ denotes the skewness of the data after applying Box-Cox. This approach is particularly suitable when iterative edits cause systematic or progressive shifts in one side of the distribution. By forcing the transformed data closer to symmetry, skewness minimization can mitigate overlap in scenarios with moderate but persistent edits.

\textbf{Strategy 3: Kurtosis Minimization.}
The third strategy seeks to diminish heavy tails by minimizing $K(\lambda)$, the kurtosis of the transformed data. High kurtosis often indicates that a handful of extreme values dominate the distribution. This situation can arise when aggressive or compound edits lead to significant reconstruction errors for particular images. Flattening the tails helps prevent outliers from overshadowing typical samples, thereby improving the overall separability of edited and unedited images.

\textbf{Suitability and Conclusion.} Our $\lambda$-selection process operates like a hyperparameter search, where each strategy (MLE, Skewness Minimization, or Kurtosis Minimization) scans potential $\lambda$ values and chooses the one that best addresses the prevalent distortion in the data. In practice, all three strategies enhance the separability of generated vs.\ edited images, but MLE typically provides the highest accuracy and stability under a broad range of editing intensities.
%
%
\subsection{Classification Mechanism}
\label{subsection:svm_classification}
To classify images, we collect the Adversarial Editing Dataset (AE-Dataset) (see \S~\ref{subsection:experiment_setup}), where the positive class consists of images generated by various generative models, and the negative class includes all manipulated images edited iteratively or manually. Each image’s reconstruction loss is first computed and transformed using the Box-Cox transformation. A simple linear Support Vector Machine (SVM) is then trained on the transformed losses, leveraging the simplicity of the one-dimensional space in which these values reside. The training process ensures that the selected $\lambda$ parameter effectively separates the generated images (positive class) from all manipulated images (negative class). For a new input image, the reconstruction loss is computed, transformed, and fed into the trained SVM, which outputs a binary decision: “generated” or “manipulated,” based on the sign of the decision function. This approach is computationally efficient while maintaining robust performance, distinguishing between generated and manipulated content.

\section{Experiment}
This section comprehensively evaluates \textsc{LambdaTracer}. 
We outline the experimental settings, including model choices, prompt design, dataset composition, baseline selection, and metrics definition (\S~\ref{subsection:experiment_setup}). 
Next, we compare \textsc{LambdaTracer} against the baseline~\cite{wang2024trace} in detecting various manipulated images (\S~\ref{subsection:performance_manipulation}). We present an ablation study (\S~\ref{subsection:ablation}) to examine the contribution of each component in \textsc{LambdaTracer} in achieving high performance. Then, we demonstrate the scalability and adaptability of our $\lambda$-selection mechanism (\S~\ref{subsection:performance_lambda_selection}). Together, these analyses underscore the effectiveness and robustness of our approach in dynamic, open-environment scenarios.
\begin{table*}[h!]
\centering
\small
\caption{Performance comparison between the baseline \textsc{LatentTracer}~\cite{wang2024trace} and our method \textsc{LambdaTracer} on manipulated images.
Manipulations include manual and iterative text-guided editing methods. For iterative editing, the same text-guided editing method is applied repeatedly on a single generated image for 1 to 5 iterations, simulating progressive modifications. Performance metrics are reported for each iteration (1–5) and for Aggregate Iteration, which combines all manipulated images from iterations 1 to 5 to evaluate overall performance across the iterative editing process. Details of the manipulated groups can be found in Appendix~\ref{appendix:manipulated_group}.}
\label{tab:iterative_editing}
\begin{tabular}{llll|lll}
\toprule
\multirow{2}{*}{Manipulated Group}& \multicolumn{3}{c}{\textsc{LatentTracer} (baseline)} & \multicolumn{3}{c}{\textsc{LambdaTracer} (ours)} \\ 
\cmidrule(lr){2-4} \cmidrule(lr){5-7}
 &Precision& Recall & F1-Score  &Precision& Recall & F1-Score \\ 
\midrule\midrule
Photoshop Modified &0.6441&
0.5429&0.5891&0.7778 \scriptsize($\uparrow$ 13.4 \%)&0.9500 \scriptsize($\uparrow$ 40.7 \%) & 0.8553 \scriptsize($\uparrow$ 26.6 \%)\\
1 Iteration &0.6070&0.8000 &0.6903 & 0.6658 \scriptsize($\uparrow$ 5.9 \%)&0.9250 \scriptsize($\uparrow$ 12.5 \%) & 0.7743 \scriptsize($\uparrow$ 8.4 \%)\\
2 Iteration &0.6292&  0.8000 & 0.7044 & 0.6602 \scriptsize($\uparrow$ 3.1 \%)&0.9714 \scriptsize($\uparrow$ 17.1 \%) & 0.7861 \scriptsize($\uparrow$ 8.2 \%)\\
3 Iteration  &0.6364& 0.8000 & 0.7089 & 0.6640 \scriptsize($\uparrow$ 2.8 \%)&0.8964 \scriptsize($\uparrow$ 9.6 \%)& 0.7629 \scriptsize($\uparrow$ 5.4 \%)\\
4 Iteration  &0.6788& 0.8000 & 0.7344 & 0.6867 \scriptsize($\uparrow$ 0.8 \%)&0.9000 \scriptsize($\uparrow$ 10.0 \%)& 0.7790 \scriptsize($\uparrow$ 4.5 \%) \\
5 Iteration  &0.6935& 0.8000 & 0.7430 & 0.6958 \scriptsize($\uparrow$ 0.2 \%)&0.8964 \scriptsize($\uparrow$ 9.6 \%) & 0.7825 \scriptsize($\uparrow$ 4.1 \%)\\
Aggregate Iteration  &0.6512& 0.8000 & 0.7179  & 0.6597 \scriptsize($\uparrow$ 0.8 \%)&0.9321 \scriptsize($\uparrow$ 13.2 \%) & 0.7726 \scriptsize($\uparrow$ 5.5 \%)\\
\bottomrule
\end{tabular}
\end{table*}

\subsection{Experiment Setup}~\label{subsection:experiment_setup}
\textbf{Models.} The experiments utilize several state-of-the-art generative and editing models, including Stable Diffusions (v1-5, v2-base, XL-1.0-base)~\cite{rombach2022high}, Kandinsky~\cite{liu2024latentguardsafetyframework}, Anytext~\cite{tuo2023anytext}, ControlNet~\cite{zhang2023adding,zhao2024uni,li2025controlnet}, and InstructPix2Pix~\cite{brooks2023instructpix2pix}. 

\textbf{Prompts.} For generative models, we select the first 20 prompts from \textsc{LatentTracer}~\cite{wang2024trace} to ensure consistency and comparability. Additionally, we designed 20 corresponding prompts specifically for text-guided editing methods~\cite{brooks2023instructpix2pix,zhang2023adding,zhao2024uni,li2025controlnet}, systematically evaluating the effects of iterative editing while focusing on diverse artistic themes and attributes. 

\textbf{Datasets.} We extend the datasets by applying modifications through Adobe Photoshop and iterative text-guided editing methods using InstructPix2Pix~\cite{brooks2023instructpix2pix} and ControlNet~\cite{zhang2023adding,zhao2024uni,li2025controlnet}, ensuring a broader and more diverse dataset for experimentation. Specifically, the Adversarial Editing Dataset (AE-Dataset) includes $18$ categories: $5$ for generative models, $2$ for manual background modification using Adobe Photoshop, and $11$ for iterative text-guided editing. For iterative processing, the same text-guided editing method is applied repeatedly on a single generated image for $1$ to $5$ iterations, simulating progressive modifications. Each category contains $20$ samples per model, with $20$ random seeds per sample, resulting in $400$ images per category. In total, our AE-Dataset comprises $7200$ images.

\textbf{Baseline.} We choose \textsc{LatentTracer}~\cite{wang2024trace} as the baseline for comparison. Unlike embedding-based methods (e.g., watermarking and model fingerprinting) that require additional steps during the training or generation phases, \textsc{LatentTracer} is the method capable of achieving alteration-free origin attribution. Moreover, \textsc{LatentTracer} has been shown to outperform its predecessor, RONAN~\cite{wang2024did}, making it a more suitable and robust benchmark for evaluating our proposed approach.

\textbf{Evaluation Metrics.} We focus on Precision, Recall, and F1-score specifically for modified images. Precision measures how many flagged images are modified, while Recall measures how many modifications are correctly detected. The F1-score balances both, providing a comprehensive assessment of detection performance. By measuring these metrics, we show that our approach enhances the robustness of manipulation detection, ensuring more reliable differentiation between original and altered content. Formulas can be found in Appendix~\ref{appendix:edited_prompts}.

\subsection{Detection Performance on Manipulated Images}\label{subsection:performance_manipulation}
To evaluate the effectiveness of our proposed method in distinguishing manipulated images, we conducted a series of experiments comparing it against the baseline \textsc{LatentTracer}\cite{wang2024trace} using AE-Dataset. Precision, Recall, and F1-Score were measured across five iterative editing steps and Photoshop-modified images, with aggregated results reported to provide an overall assessment.

Table \ref{tab:iterative_editing} illustrates that our \textsc{LambdaTracer} consistently outperforms the baseline in both Recall and F1-Score across all manipulations. 
Specifically, the Recall scores demonstrate that \textsc{LambdaTracer} is more effective in detecting actual modifications, which is crucial for identifying malicious users attempting to manipulate content. The F1-Scores further confirm the balanced improvement in precision and recall, underscoring the robustness of \textsc{LambdaTracer} in maintaining high attribution accuracy. Notably, our ablation study underscores the Box-Cox transformation’s role in enhancing latent space separability, leading to improved detection performance.

These results highlight the superiority of \textsc{LambdaTracer} in accurately distinguishing various types of modified content, thereby enhancing the reliability of attribution tasks. Its ability to maintain high performance across multiple editing iterations and manual Photoshop modifications demonstrates \textsc{LambdaTracer}’s adaptability and effectiveness in real-world open environments, where content manipulation may occur repeatedly or involve diverse techniques.
\begin{table}[h!]
\centering
\small
\caption{Ablation Study of \textsc{LambdaTracer} conducted using AE-Dataset, where the positive class includes images generated by Stable Diffusion v2-base, v1-5, XL-1.0-base, and Kandinsky models, and the negative class includes manipulated images edited using InstructPix2Pix and ControlNet. We compare the performance of our \textsc{LambdaTracer} against various ablated versions and \textsc{LatentTracer}~\cite{wang2024trace}, highlighting the impact of the newly designed Box-Cox transformation on performance.}
\label{tab:ablation}
\begin{tabular}{lllll}
\toprule
Method &Precision& Recall & F1-Score \\ 
\midrule\midrule
\textsc{LambdaTracer} &\textbf{0.6636}& \textbf{0.8839} & \textbf{0.7581} \\
\textsc{Lambda} \scriptsize(w/o box-cox)  &0.6610& 0.6304 & 0.6453 \\\midrule
\textsc{Lambda} \scriptsize(w/ logarithmic) &0.6604& 0.6321 & 0.6459 \\
\textsc{Lambda} \scriptsize(w/ power) &0.6190& 0.5071 & 0.5575 \\
\textsc{Lambda} \scriptsize(w/ exponential)  &0.6623& 0.6304 & 0.6460 \\
\textsc{Lambda} \scriptsize(w/ z-score) &0.6301& 0.6571 & 0.6433 \\\midrule
\textsc{LatentTracer} &0.6447& 0.5161 & 0.5733 \\
\textsc{Latent} \scriptsize(w/ box-cox) &0.6364& 0.5875 & 0.6110 \\
\bottomrule
\end{tabular}
\end{table}
\begin{figure*}[h]
    \centering
    \subfigure[Precision]{ \includegraphics[width=0.32\linewidth]{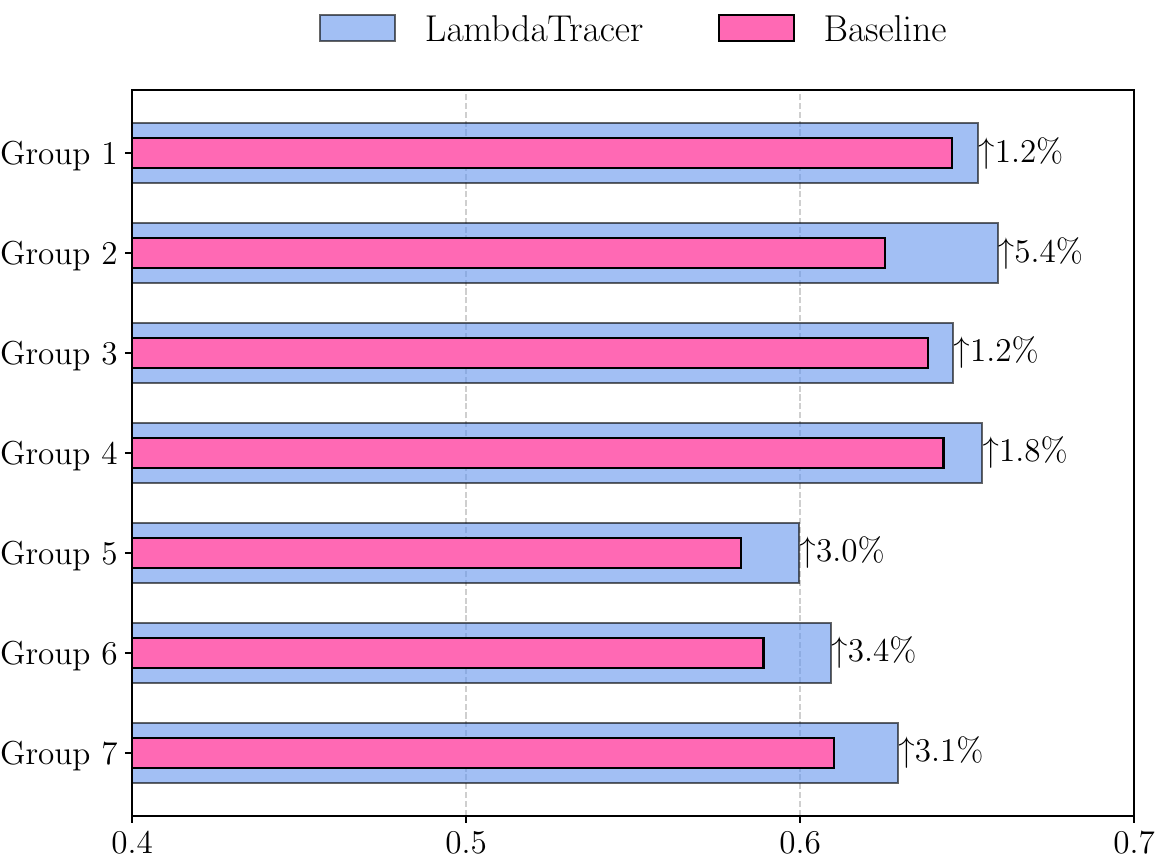}
    }
    \subfigure[Recall]{\includegraphics[width=0.32\linewidth]{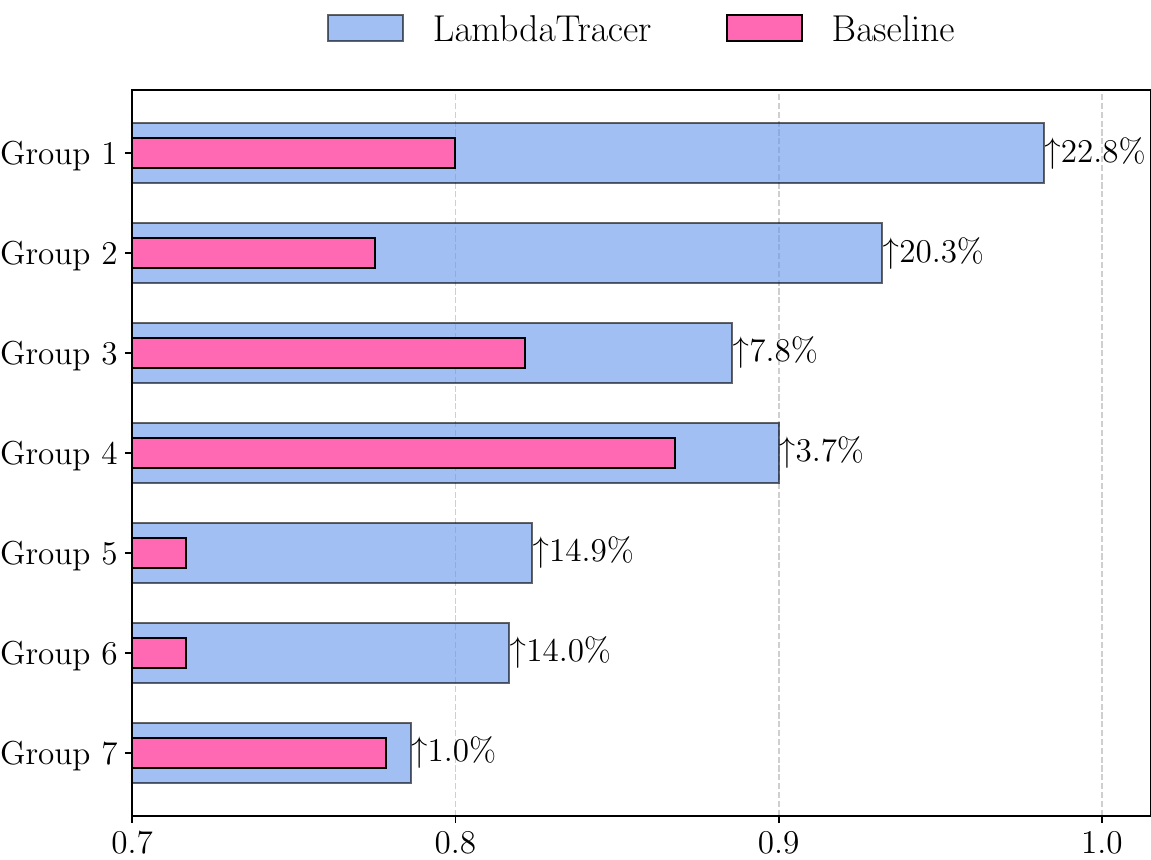}
    }
    \subfigure[F1-Score]{
    \includegraphics[width=0.32\linewidth]{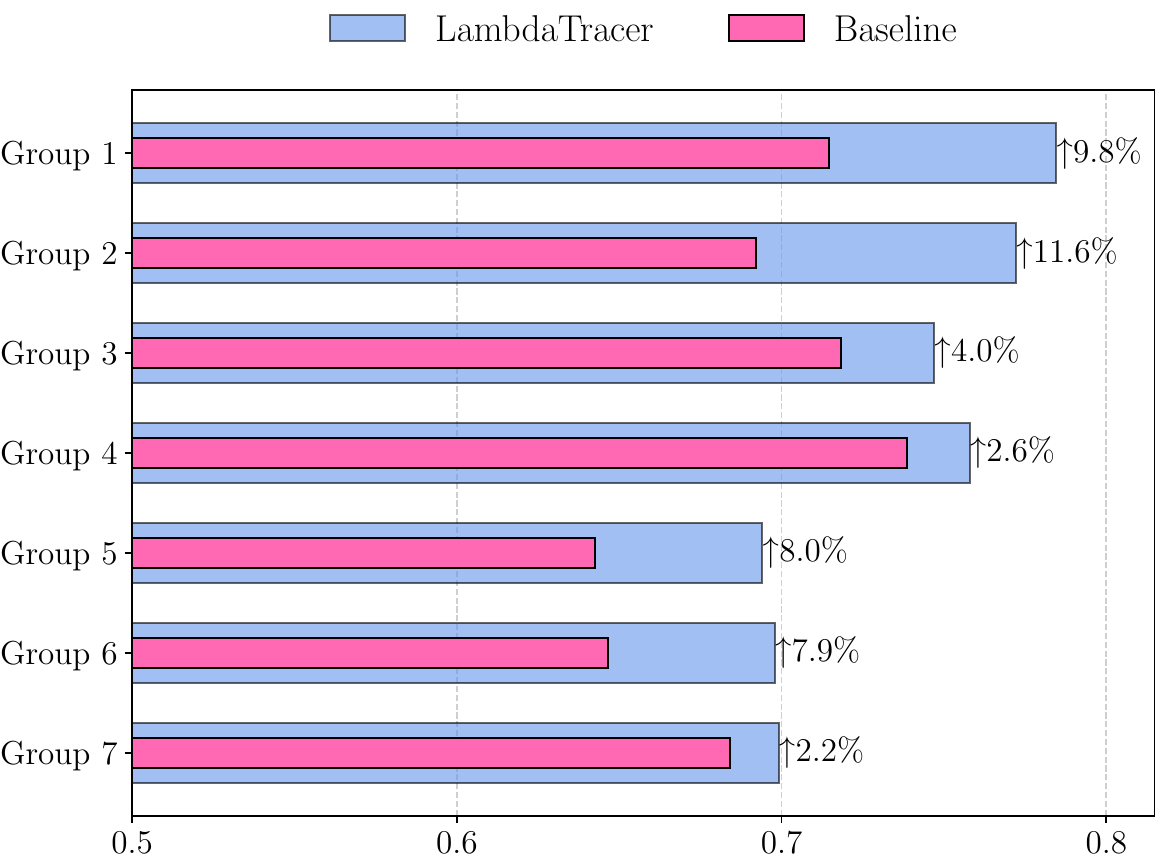}
    }
    \vspace{-10pt}
    \caption{Comparison of Baseline and \textsc{LambdaTracer} on manipulated images for different positive groups. 
    Each group comprises various generative models, while the negative class covers various manipulations (e.g., Photoshop-based edits and iterative text-guided modifications). 
    \textsc{LambdaTracer}, equipped with its $\lambda$-selection strategy, demonstrates consistent improvements over the Baseline}
    \label{fig:precision_recall_f1score}
    \vspace{-10pt}
\end{figure*}

\subsection{Analysis of Ablation Studies}\label{subsection:ablation}
\textbf{Effectiveness of Box-Cox Transformation.} To validate the effectiveness of the proposed Box-Cox transformation, we conducted experiments by removing the Box-Cox transformation and using only a simplified supervised learning version of \textsc{LambdaTracer}. As shown in Table \ref{tab:ablation}, the F1-Score dropped from 0.7581 to 0.6453 without the Box-Cox transformation. 

\textbf{Comparison with Alternative Transformations.} To further assess the effectiveness of Box-Cox compared to other transformations, we replaced it with four widely used alternatives: logarithmic, power, exponential, and z-score transformations. 
As shown in Table \ref{tab:ablation}, \textsc{LambdaTracer} with Box-Cox significantly outperformed other variants. 
These findings underscore the unique ability of the Box-Cox transformation to optimize the separability of generated and manipulated images, making it the most effective preprocessing method for our task.

\textbf{Performance with Baseline and Box-Cox.} Lastly, to evaluate whether the performance gains are solely attributable to the Box-Cox transformation, we applied it to the baseline model, \textsc{LatentTracer}. Although \textsc{LatentTracer} with Box-Cox slightly improved F1-Score to 0.6110, it still lagged significantly behind \textsc{LambdaTracer}. \textsc{LambdaTracer} outperformed this variant by 14.71\% 

\subsection{Adaptive $\lambda$-Selection for Generalized Performance}\label{subsection:performance_lambda_selection}
To accommodate the rapid emergence of new generative models, we redefined the composition of the positive class in this study. Specifically, we form seven positive groups to evaluate how different combinations of generative models influence detection performance (detailed in Appendix~\ref{appendix:models_and_groups}). This redesign aims to enhance the generalizability of the proposed method across diverse generative sources. Figure~\ref{fig:precision_recall_f1score} compares the performance of the baseline~\cite{wang2024trace} and our \textsc{LambdaTracer} across seven positive groups. \textsc{LambdaTracer} consistently outperforms the baseline in all three metrics, demonstrating its robustness in accurately distinguishing manipulated content.

Table~\ref{tab:overall_acc} further reports the binary classification accuracy of both methods. Across all seven positive groups, \textsc{LambdaTracer} achieves higher accuracy, showcasing the effectiveness of its adaptive $\lambda$-selection mechanism in handling variations across multiple generative models and manipulation types. These findings affirm that \textsc{LambdaTracer} not only excels in detecting manipulated images but also maintains strong scalability and generalization, making it well-suited for future open-environment scenarios with continuously evolving generative technologies.
\begin{table}[h!]
\small
\centering
\caption{Overall Accuracy Comparison of Baseline~\cite{wang2024trace} and \textsc{LambdaTracer} Across Different Positive Groups. 
Each positive group comprises diverse generative models (The definition of each group is given in  Appendix~\ref{appendix:models_and_groups}). The evaluation includes both positive and manipulated images. The results underscore that \textsc{LambdaTracer} consistently outperforms the Baseline by adaptively selecting the optimal $\lambda$.}
\label{tab:overall_acc}
 \scalebox{1.0}{
\begin{tabular}{lcc}
\toprule
Positive Class & \textsc{LatentTracer}& \textsc{LambdaTracer} (ours) \\ 
\midrule\midrule
Group~\MakeUppercase{\romannumeral 1} & 0.6804 & 0.7304 \scriptsize($\uparrow$ 5.00 \%) \\
Group~\MakeUppercase{\romannumeral 2} & 0.6554 & 0.7250 \scriptsize($\uparrow$ 6.96 \%) \\
Group~\MakeUppercase{\romannumeral 3} & 0.6807 & 0.7000 \scriptsize($\uparrow$ 1.93 \%) \\
Group~\MakeUppercase{\romannumeral 4} & 0.6929 & 0.7125 \scriptsize($\uparrow$ 1.96 \%)\\
Group~\MakeUppercase{\romannumeral 5} & 0.6012 & 0.6369 \scriptsize($\uparrow$ 3.57 \%) \\
Group~\MakeUppercase{\romannumeral 6} & 0.6083 & 0.6464 \scriptsize($\uparrow$ 3.81 \%) \\
Group~\MakeUppercase{\romannumeral 7} & 0.6305 & 0.6393 \scriptsize($\uparrow$ 0.88 \%) \\
\bottomrule
\end{tabular}
}
\end{table}

\section{Conclusion}
In this paper, we introduced \textsc{LambdaTracer}, a novel inversion-based origin attribution method designed to predict whether an image is a forged original in open and adversarial environments. By employing a newly designed Box-Cox transformation to optimize the loss function, \textsc{LambdaTracer} enhances the model’s ability to classify images accurately, distinguishing genuine originals from those modified by advanced text-guided editing models. Experimental results demonstrate that \textsc{LambdaTracer} outperforms existing methods, achieving high accuracy in detecting forged images even after iterative editing. Specifically, it effectively identifies images modified by sophisticated editing tools such as InstructPix2Pix and ControlNet, providing reliable provenance tracing for model-generated content. 
This work offers a robust solution for copyright protection and the responsible commercialization of diffusion models. Future work includes enhancing the interpretability of \textsc{LambdaTracer} through visualization techniques that provide intuitive insights into its classification decisions.

\newpage
\section*{Impact Statement}
The primary purpose of this work is to combat copyright infringement and unauthorized modifications of model-generated content by providing robust origin attribution capabilities. \textsc{LambdaTracer} is designed to ensure the responsible use of generative models, particularly in creative domains, by tracing and identifying misuse. This technology is not intended for use by adversarial actors or those seeking to exploit generative models for unlawful purposes. We encourage its application within ethical and legal frameworks to safeguard intellectual property while promoting responsible artificial intelligence practices.



\nocite{*}
\bibliography{paper}
\bibliographystyle{icml2025}

\newpage
\appendix
\onecolumn
\section{Methodology: Formula}~\label{appendix:formula}
\textbf{Strategies of $\lambda$ Selection.} $\lambda_{\text{Skew}}$ is $arg\min_\lambda \left| \text{Skew} \left( T_\lambda(\text{data}) \right) \right|$. Equation~\ref{equ:skew} illustrates the skewness of the transformed data. $\lambda_{\text{Kurt}}$ is $\arg\min_\lambda |\text{Kurtosis} \left( T_\lambda(\text{data}) \right) - c|$, where $c$ is the constant to adjust the kurtosis. Equation~\ref{equ:kurtosis} illustrates the kurtosis of the transformed data.

\begin{align}\label{equ:skew}
\text{Skew} \left( T_\lambda(\text{data}) \right) = \frac{1}{n} \sum_{i=1}^n \left( \frac{T_\lambda(x_i) - \mu_\lambda}{\sigma_\lambda} \right)^3,
\end{align}
where $T_\lambda(x_i)$ denotes the value of the data point $x_i$ after applying the Box-Cox transformation with current temporary $\lambda$ (i.e. Equation~\ref{equ:boxcox}), $\mu_\lambda$ denotes the mean of the transformed data $T_\lambda(\text{data})$, $\sigma_\lambda$ denotes the standard deviation of the transformed data $T_\lambda(\text{data})$, and $n$ is the total number of data points.
\begin{align} ~\label{equ:kurtosis}
\text{Kurtosis} \left( T_\lambda(\text{data}) \right) = \frac{1}{n} \sum_{i=1}^n \left( \frac{T_\lambda(x_i) - \mu_\lambda}{\sigma_\lambda} \right)^4
\end{align}
where $T_\lambda(x_i)$ denotes the value of the data point $x_i$ after applying the Box-Cox transformation with current temporary $\lambda$ (i.e. Equation~\ref{equ:boxcox}), $\mu_\lambda$ denotes the mean of the transformed data $T_\lambda(\text{data})$, $\sigma_\lambda$ denotes the standard deviation of the transformed data $T_\lambda(\text{data})$, and $n$ is the total number of data points.

\section{Methodology: Additional Transformations}\label{appendix:transformations}
\textbf{Z-score Standardization} is a widely used linear transformation that scales data to have a mean of zero and a standard deviation of one, ensuring uniformity and computational efficiency~\cite{vapnik1996support,cervantes2020comprehensive}. However, its linear nature is insufficient for distinguishing overlapping distributions with subtle nonlinear variations, as it fails to capture complex patterns in the data. This limitation highlights the need for nonlinear transformations to enhance separability in such scenarios. 

\textbf{Logarithmic Transformation} is adequate for handling right-skewed data by compressing large values and stabilizing variance, reducing the influence of outliers. However, their fixed transformation intensity limits their ability to amplify minor differences in overlapping regions, making them less suitable for complex distributions requiring adaptive separability enhancements. 

\textbf{Exponential Transformation} effectively reduces left-skewness by amplifying smaller values, enhancing separability in specific cases. However, they are unsuitable for our right-skewed data, as they exacerbate skewness and distort the intrinsic structure, compromising essential distinguishing features. 

\textbf{Power Transformation} generalizes logarithmic and exponential methods by applying flexible powers to data, effectively reducing skewness and stabilizing variance across diverse distributions. While they retain essential data characteristics, their fixed parameters limit adaptability to varying distributions, leading to suboptimal separability in complex, overlapping datasets.

\section{Algorithm: Pseudo-code}~\label{appendix:algo}
Algorithms~\ref{alg:kurtosis_min} and \ref{alg:skew_min} illustrate two strategies for applying the Box-Cox transformation. The first strategy, Kurtosis Minimization, selects the parameter $\lambda$ by minimizing the distribution's kurtosis. In contrast, the second strategy, Skewness Minimization, focuses on reducing the skewness of the transformed data to improve distribution symmetry.

\begin{figure}[h]
\centering
\begin{minipage}{0.45\textwidth}
\begin{algorithm}[H]
\caption{Kurtosis Minimization for $\lambda$ Selection}
\label{alg:kurtosis_min}
\textbf{Input:} Loss Values $x$ \\
\textbf{Output:} Transformed Data $t$, Optimal Parameter $\lambda^*$
\begin{algorithmic}[1]
\Function{BoxCox-Kurtosis}{$x$}
    \State $x \gets \{x_i + 10^{-8} \mid x_i \in x\}$
    \State $\Lambda \gets \{\lambda_1, \lambda_2, \dots, \lambda_n\}$
    \State $\lambda^* \gets \text{None}$
    \State $\mathcal{K}_{min} \gets \infty$
    \For{$\lambda \in \Lambda$}
        \State $z \gets \text{BoxCox}(x, \lambda)$
        \State $\text{current} \gets |\text{Kurtosis}(z) - 1|$
        \If{$\text{current} < \mathcal{K}_{min}$}
            \State $\mathcal{K}_{min} \gets \text{current}$
            \State $\lambda^* \gets \lambda$
        \EndIf  
    \EndFor     
    \State $t \gets \text{BoxCox}(x, \lambda^*)$
    \State \Return $t, \lambda^*$
\EndFunction    
\end{algorithmic}
\end{algorithm}
\end{minipage}
\hfill
\begin{minipage}{0.45\textwidth}
\begin{algorithm}[H]
\caption{Skewness Minimization for $\lambda$ Selection}
\label{alg:skew_min}
\textbf{Input:} Loss Values $x$ \\
\textbf{Output:} Transformed Data $t$, Optimal Parameter $\lambda^*$
\begin{algorithmic}[1]
\Function{BoxCox-Skewness}{$x$}
    \State $x \gets \{x_i + 10^{-8} \mid x_i \in x\}$
    \State $\Lambda \gets \{\lambda_1, \lambda_2, \dots, \lambda_n\}$
    \State $\lambda^* \gets \text{None}$
    \State $\mathcal{S}_{min} \gets \infty$
    \For{$\lambda \in \Lambda$}
        \State $z \gets \text{BoxCox}(x, \lambda)$
        \State $\text{current} \gets |\text{Skewness}(z) - 1|$
        \If{$\text{current} < \mathcal{S}_{min}$}
            \State $\mathcal{S}_{min} \gets \text{current}$
            \State $\lambda^* \gets \lambda$
        \EndIf  
    \EndFor     
    \State $t \gets \text{BoxCox}(x, \lambda^*)$
    \State \Return $t, \lambda^*$
\EndFunction    
\end{algorithmic}
\end{algorithm}
\end{minipage}
\end{figure} 

\section{Experiment Setup}\label{appendix:edited_prompts}
\textbf{Environments.} The experiments are implemented with Python 3.10 and PyTorch 2.0 on a system equipped with NVIDIA A6000 GPUs running Ubuntu 20.04.

\textbf{Prompts.} Table~\ref{tab:appendix_edit_prompts} illustrates that prompts are designed to elicit diverse modifications, ranging from artistic refinements to structural alterations, showcasing the versatility and adaptability of the models. This detailed list complements the results presented in the main paper, enabling reproducibility and facilitating further exploration of text-guided editing capabilities.
\begin{table*}[h]
\centering
\caption{A list of 20 prompts used for text-guided editing methods, including ControlNet and InstructPix2Pix. These prompts were carefully designed to evaluate the capabilities of the editing models across diverse artistic and structural modifications.}
\label{tab:appendix_edit_prompts}
\begin{tabular}{ll}
\toprule
Prompt 1 & ``Add a soft glow to the edges of the background lights." \\ \midrule
Prompt 2 & ``Enhance the brightness of the beach sand slightly, keeping everything else unchanged." \\ \midrule
Prompt 3 & ``Add a subtle warm hue to the lighting around the woman's face." \\ \midrule
Prompt 4 & ``Add a faint glow to the largest mushroom in the background." \\ \midrule
Prompt 5 & ``Enhance the stained glass colours slightly for better contrast." \\ \midrule
Prompt 6 & ``Add a faint shadow of a tree in the park background." \\ \midrule
Prompt 7 & ``Slightly enhance the brightness of the glowing ball in her hands." \\ \midrule
Prompt 8 & ``Add faint reflections on the wet street in the background." \\ \midrule
Prompt 9 & ``Introduce a slight shadow effect under the gym equipment in the background." \\ \midrule
Prompt 10 & ``Add a faint mist to the night city background, keeping the figure untouched." \\ \midrule
Prompt 11 & ``Add a slight ripple effect to the water on the beach background." \\ \midrule
Prompt 12 & ``Add subtle smoke effects near the dragon's claws in the distance." \\ \midrule
Prompt 13 & ``Introduce a faint reflection on the wet surface in the cyberpunk city background." \\ \midrule
Prompt 14 & ``Add a slight glow to the top of the tallest skyscraper." \\ \midrule
Prompt 15 & ``Enhance the reflections on the wet street in the background slightly." \\ \midrule
Prompt 16 & ``Add a faint flicker effect to the candles in their hands." \\ \midrule
Prompt 17 & ``Add subtle smoke near the Tigrex’s feet." \\ \midrule
Prompt 18 & ``Introduce a faint shadow behind the joker in the background." \\ \midrule
Prompt 19 & ``Add a subtle glow to the raindrops in the background." \\ \midrule
Prompt 20 & ``Add faint footprints in the sand near the cat." \\ 
\bottomrule
\end{tabular}
\end{table*}

\textbf{Evaluation Metrics.} Precision (Eq.~\ref{eq:precision}) indicates the proportion of flagged images that are genuinely modified. A high Precision value helps maintain the system’s trustworthiness by minimizing false positives. Recall (Eq.~\ref{eq:recall}) captures the proportion of actual modified images correctly identified by the model. A high Recall ensures malicious or tampered images do not go undetected. F1-score (Eq.~\ref{eq:f1score}) offers a balanced measure by harmonically combining Precision and Recall, particularly useful when false positives and negatives have significant consequences.
\begin{equation}
\label{eq:precision}
\text{Precision} = \frac{TP}{TP + FP}
\end{equation}
\begin{equation}
\label{eq:recall}
\text{Recall} = \frac{TP}{TP + FN}
\end{equation}
\begin{equation}
\label{eq:f1score}
F1\text{-score} = 2 \times \frac{\text{Precision} \times \text{Recall}}{\text{Precision} + \text{Recall}}
\end{equation}
\noindent
Here, TP (True Positives) are the correctly identified modified images, FP (False Positives) are the mistakenly flagged images, and FN (False Negatives) are the missed modified images. By centring our evaluation on these measures, we address the most critical aspect of content integrity—accurate detection of manipulations—thus improving security and trust within the publishing ecosystem.

\begin{figure}[h]
    \centering
    \includegraphics[width=1\textwidth]{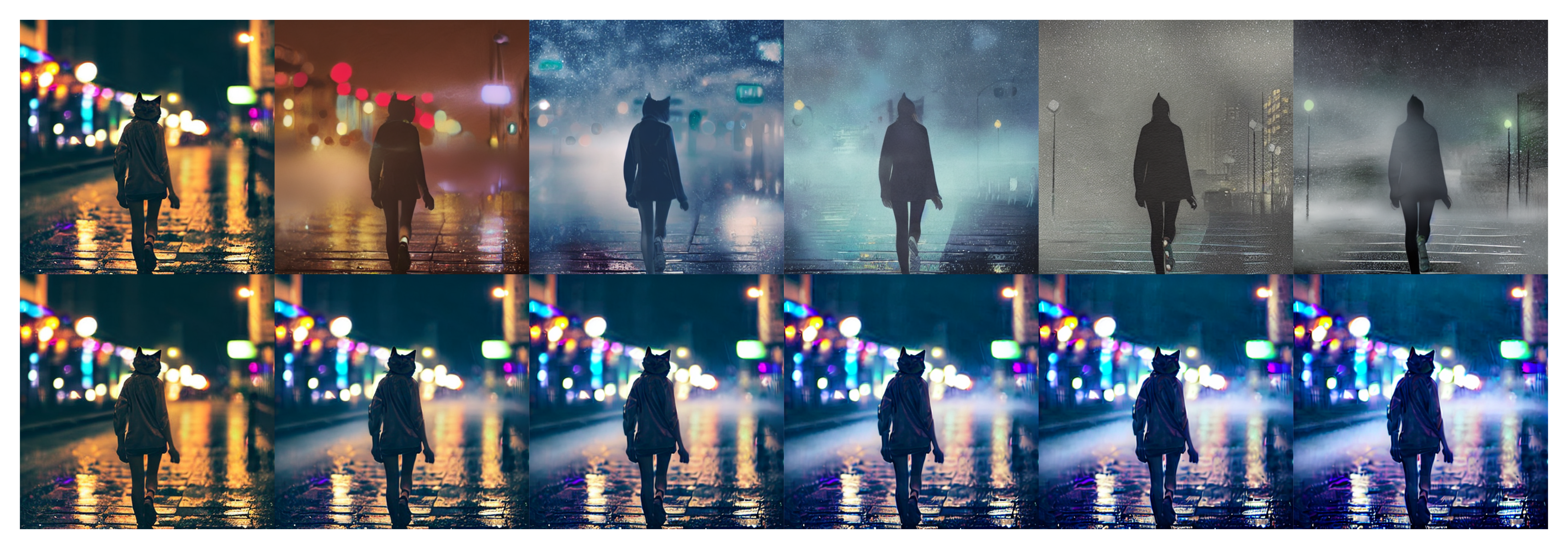}
    \vspace{-20pt} 
    \caption{Example of iterative text-guided editing on images generated by SD v2-base. The top row showcases images edited progressively using ControlNet, with iterations ranging from 1 to 5. The bottom row demonstrates images modified using InstructPix2Pix, also iterated from 1 to 5 times. This visualization highlights the cumulative effects of iterative modifications introduced by different editing models.}
    \label{fig:appendix_iterative_editing_examples}
\end{figure}

\textbf{Datasets.}
We compile a dataset of 18 categories, each containing 400 images at a resolution of \(512 \times 512\). These categories encompass images generated by the models detailed in Appendix~\ref{appendix:models_and_groups} and images further modified via Adobe Photoshop, InstructPix2Pix, and ControlNet. This initial dataset provides a balanced scope of generated and edited content, ensuring comprehensive coverage of common manipulation types. (We plan to expand each category with additional images in future work, contributing to our community by providing diverse and representative adversarial datasets.) Fig.~\ref{fig:appendix_iterative_editing_examples} illustrates the progressive modifications introduced by text-guided editing tools on images originally generated by
Stable Diffusion v2-base. The top row demonstrates the cumulative effects of iterative editing using ControlNet, while the bottom row
showcases similar results with InstructPix2Pix. These tools preserve the original images' artistic essence and introduce subtle yet impactful changes with each iteration. Such capabilities highlight the strength of modern
text-guided editing models in creating visually appealing modifications while simultaneously posing significant challenges
for detection and provenance tracing. Notably, this figure is derived from Fig.~\ref{fig:6imgs}, extending the analysis to include more
iterations and exploring the compounding effects of iterative editing.

\section{Details of Manipulated Group Compositions}\label{appendix:manipulated_group}
\textbf{Positive Group Compositions.}
We form seven manipulated groups to evaluate the performance under different manipulations. This setup is designed to test the method's robustness, ensuring it maintains strong performance even after multiple iterations of text-guided editing. Specifically:
\begin{itemize}
    \vspace{-5pt} 
    \item \textbf{Photoshop Modified:} Images generated by generative models (Stable Diffusion v2-base, v1-5) are manually edited using Adobe Photoshop\footnote{https://www.adobe.com/ca/products/photoshop.html}. Specifically, the background of each image is replaced with a different colour.\vspace{-2pt} 
    
    \item \textbf{1 Iteration:} Images generated by generative models (Stable Diffusion v2-base, v1-5) are edited once using text-guided editing methods. This includes (1) applying InstructPix2Pix to modify image content based on textual instructions and (2) using ControlNet to alter specific aspects of the images according to control prompts.\vspace{-2pt} 
    
    \item \textbf{2 Iteration:} Images generated by generative models (Stable Diffusion v2-base, v1-5) are subjected to two consecutive rounds of editing using the same text-guided editing method. Specifically, InstructPix2Pix or ControlNet is applied twice on the same generated image, where the first edit's output serves as the second edit's input. This iterative process allows for progressive modifications to the image, simulating more complex or cumulative changes. \vspace{-2pt} 
    
    \item \textbf{3 Iteration:} Images generated by generative models (Stable Diffusion v2-base, v1-5) are subjected to three consecutive rounds of editing using the same text-guided editing method. Specifically, InstructPix2Pix or ControlNet is applied thrice on the same generated image, where the output of the second edit serves as the input for the third edit.\vspace{-2pt} 
    
    \item \textbf{4 Iteration:} Images generated by generative models (Stable Diffusion v2-base, v1-5) are subjected to four consecutive rounds of editing using the same text-guided editing method. Specifically, InstructPix2Pix or ControlNet is applied quartic on the same generated image, where the output of the third edit serves as the input for the fourth edit.\vspace{-2pt} 
    
    \item \textbf{5 Iteration:} Images generated by generative models (Stable Diffusion v2-base, v1-5) are subjected to five consecutive rounds of editing using the same text-guided editing method. Specifically, InstructPix2Pix or ControlNet is applied fifthly on the same generated image, where the output of the fourth edit serves as the input for the fifth edit.\vspace{-2pt} 
    
     \item \textbf{Aggregate Iteration:} This category combines all images edited using text-guided editing methods (InstructPix2Pix and ControlNet) across 1 to 5 iterations. In addition to individual iterative edits (e.g., applying the same method multiple times), it includes combined editing scenarios where different methods are applied sequentially. Specifically, this involves cases such as (1) using InstructPix2Pix to modify generated images followed by further edits using ControlNet, and (2) using ControlNet first to modify generated images, then applying InstructPix2Pix. By incorporating these combined editing cases, this category provides a comprehensive evaluation of overall performance across diverse editing pipelines.\vspace{-2pt}
\end{itemize}

\section{Details of Generative Models and Positive Group Compositions}
\label{appendix:models_and_groups}
\textbf{Generative Models.} We use five popular and well-developed generative models representing a diverse range of latent diffusion–based architectures. Specifically:
\begin{itemize}
\vspace{-5pt} 
\item \textbf{Stable Diffusion v2-base\footnote{https://huggingface.co/stabilityai/stable-diffusion-2-base}:}
This model employs a latent diffusion approach with a VAE for image encoding and decoding, trained at a resolution of $512 \times 512$. Compared to earlier Stable Diffusion v1 variants, it incorporates an enhanced text encoder and refined U-Net backbone for improved prompt adherence.\vspace{-2pt} 

\item \textbf{Stable Diffusion v1-5\footnote{https://huggingface.co/stabilityai/stable-diffusion-1-5}:}
This model refines the Stable Diffusion v1 series by extending fine-tuning steps at $512 \times 512$ resolution. It uses a U-Net–based diffusion process with a VAE, enabling effective text-to-image generation while preserving visual consistency and fidelity.\vspace{-2pt} 

\item \textbf{Stable Diffusion XL-1.0-base\footnote{https://huggingface.co/stabilityai/stable-diffusion-xl-base-1.0}:}
This model scales the latent diffusion framework to accommodate larger training data and higher capacity. It is initially trained at $256 \times 256$ for $600{,}000$ steps, then further trained at $512 \times 512$ for $200{,}000$ steps, which improves image quality and flexibility in handling varied aspect ratios.\vspace{-2pt} 

\item \textbf{Kandinsky 2-1\footnote{https://huggingface.co/spaces/ai-forever/Kandinsky2.1}:}
This model integrates a latent diffusion architecture with advanced text encoders and cross-attention mechanisms. It is trained on a large text-image dataset, allowing for a broad range of generated styles and content without compromising image coherence.\vspace{-2pt} 

\item \textbf{AnyText v1-1-2\footnote{https://huggingface.co/spaces/modelscope/AnyText}:}
This model adopts a customized text-to-image pipeline built on a latent diffusion backbone, incorporating specialized domain data for flexible and adaptive image synthesis. It aims to balance prompt fidelity with diverse generation capabilities.\vspace{-2pt} 
\end{itemize}

\textbf{Positive Group Compositions.}
We form seven positive groups (Groups~1--7) to evaluate how different combinations of the above models influence detection performance. Specifically:
\begin{itemize}
    \vspace{-5pt} 
    \item \textbf{Group~\MakeUppercase{\romannumeral 1}:} Stable Diffusion v2-base and Stable Diffusion v1-5.\vspace{-2pt} 
    \item \textbf{Group~\MakeUppercase{\romannumeral 2}:} Stable Diffusion v2-base and Kandinsky 2-1.\vspace{-2pt} 
    \item \textbf{Group~\MakeUppercase{\romannumeral 3}:} Stable Diffusion v2-base and Stable Diffusion XL-1.0-base.\vspace{-2pt} 
    \item \textbf{Group~\MakeUppercase{\romannumeral 4}:} Stable Diffusion v2-base and AnyText v1-1-2.\vspace{-2pt} 
    \item \textbf{Group~\MakeUppercase{\romannumeral 5}:} Stable Diffusion v2-base, Stable Diffusion v1-5, and Kandinsky 2-1.\vspace{-2pt} 
    \item \textbf{Group~\MakeUppercase{\romannumeral 6}:} Stable Diffusion v2-base, Stable Diffusion v1-5, and Stable Diffusion XL-1.0-base.\vspace{-2pt} 
    \item \textbf{Group~\MakeUppercase{\romannumeral 7}:} Stable Diffusion v2-base, Stable Diffusion v1-5, and AnyText v1-1-2.\vspace{-2pt} 
\end{itemize}

\section{Case Studies Highlighting Methodological Advantages}
\textbf{InstructPix2Pix-Manipulated Images vs. Stable Diffusion v1-5 Generated Images.} We evaluate the classification task between images manipulated using InstructPix2Pix with varying iteration counts (e.g., one, two, …, five iterations) and images generated by Stable Diffusion v1-5, as described in Appendix~\ref{appendix:models_and_groups}. The proposed method, \textsc{LambdaTracer}, consistently outperforms the baseline, \textsc{LatentTracer}, across all iteration levels. The improvement in this scenario is particularly noteworthy, as the baseline exhibits significant difficulty due to the highest degree of overlap in reconstructed loss distributions between these two categories (as shown in Figure~\ref{fig:heatmap_ori}). This overlap renders this case the most challenging for the baseline, underscoring the enhanced capability of \textsc{LambdaTracer} in addressing such complex classification tasks.

\textbf{Photoshop-Manipulated Images vs. Model-Generated Images}. We further assess the classification task between images manipulated using Adobe Photoshop and images generated by various models. Across all tested scenarios, \textsc{LambdaTracer} demonstrates superior performance compared to the baseline, \textsc{LatentTracer}. The improvement in this case is substantial. As shown in Table~\ref{tab:iterative_editing}, when combining images generated by multiple models with Photoshop-manipulated images for classification, \textsc{LambdaTracer} achieves a 26.6\% increase in F1-score. This result highlights the robustness and effectiveness of the proposed approach in reliably distinguishing between these categories.

\end{document}